\documentclass[11pt]{article}

\usepackage[final]{acl}

\usepackage{times}
\usepackage{latexsym}
\usepackage[T1]{fontenc}
\usepackage[utf8]{inputenc}
\usepackage{microtype}
\usepackage{inconsolata}
\usepackage{graphicx}
\usepackage{amsmath}
\usepackage{amssymb}
\usepackage{booktabs}
\usepackage{xspace}

\newcommand{\savings}{\ensuremath{\mathcal{S}}\xspace}
\newcommand{\resurf}{\ensuremath{R}\xspace}
\newcommand{\probe}{\ensuremath{A_{\mathrm{probe}}}\xspace}

\title{Suppressed, Not Erased: A Representational Trace of Edited Facts\\
Survives Even Weight-Free Knowledge Editing}

\author{Priyansh Srivastava \\
  Sirena Ai \\
  \texttt{priyansh@sirenatech.com} \\\And
  Romit Chatterjee \\
  Independent Researcher \\
  \texttt{chatterjeeromit86@gmail.com}}

\begin{document}
\maketitle

\begin{abstract}
Knowledge-editing benchmarks certify \emph{local correctness}---whether an edited model produces the new fact on near-edit prompts---but not how much of the original fact remains decodable inside the model. We study residual knowledge directly with a linear \emph{trace probe}: after editing a fact, we ask whether the original object is still recoverable from the model's hidden states. On GPT-2-XL, across three mechanistically distinct editors applied to 50 CounterFact edits, the original object remains linearly decodable well above chance after a successful edit (probe accuracy $0.96$ for ROME, $0.86$ for constrained fine-tuning, and $0.79$ for the memory-based editor GRACE, against a chance level of $0.50$; all edits reach $100\%$ generation-based success). The GRACE result is the most informative: GRACE changes \emph{zero} base-model weights, overriding the fact through an external memory, yet the original object is still decodable from the underlying network---so the residual trace cannot be attributed to an incomplete weight update. We read this as evidence that editing, even when behaviorally successful, suppresses rather than erases the original association in representational space. We also report a relearning-savings instrument that did not behave reliably in our setting and discuss why; we treat it as a negative methodological result rather than evidence. Code and data are released.
\end{abstract}

\section{Introduction}
\label{sec:intro}

Parametric knowledge editing---changing an individual fact in a trained language model without retraining---has become a standard tool, with locate-then-edit methods such as ROME \citep{meng2022locating} and MEMIT \citep{meng2023memit}, constrained fine-tuning \citep{zhu2020modifying}, and memory-based editors such as GRACE \citep{hartvigsen2023aging}. The field's evaluation, however, largely asks a single question: does the model now produce the \emph{new} fact near the edit? Standard metrics---efficacy, paraphrase generalization, neighborhood specificity \citep{meng2022locating}---all certify the presence of the new fact, and recent audits show even that certificate is fragile under realistic, generation-based evaluation \citep{yang2025mirage,cohen2024ripple}.

A different question is whether the \emph{original} fact is truly gone. A transformer does not store a fact at an address that can be cleared; a fact is enacted by distributed computation---feed-forward key--value associations \citep{geva2021transformer}, approximately linear relational decodings \citep{hernandez2024linearity}---that an edit can reshape but need not eliminate. If editing reshapes rather than deletes, then a successful edit should be compatible with the original object remaining \emph{decodable} from internal states. Prior extraction work supports this possibility: \citet{patil2024can} recovered ``deleted'' answers from ROME-edited GPT-J about $38\%$ of the time using hidden-state and rephrasing attacks.

We test this directly with a simple, controlled instrument: a linear \emph{trace probe} that attempts to recover the original object from the edited model's hidden states (\S\ref{sec:probe}). Our contribution is empirical and deliberately narrow:

\begin{itemize}
    \item We show that on GPT-2-XL, after a $100\%$-successful single edit, the original object remains linearly decodable well above chance across three mechanistically distinct editors (\S\ref{sec:results}).
    \item We isolate the \emph{mechanism} of the residual trace using GRACE \citep{hartvigsen2023aging}, a memory-based editor that changes no base weights. The original object remains decodable even under GRACE, showing the trace is not an artifact of an incomplete weight update but a property of the underlying representation the edit leaves intact (\S\ref{sec:results}).
    \item We report, transparently, a relearning-savings instrument that did not behave reliably in our setting, and analyze why memory-based editing in particular breaks its assumptions (\S\ref{sec:savings-null}). We release the full pipeline so the result is reproducible.
\end{itemize}

Our scope is one canonical editing testbed (GPT-2-XL) and a single-edit residual-trace audit. We are explicit about this throughout, and \S\ref{sec:limitations} states what the result does and does not license.

\begin{figure}[t]
  \centering
  \includegraphics[width=\columnwidth]{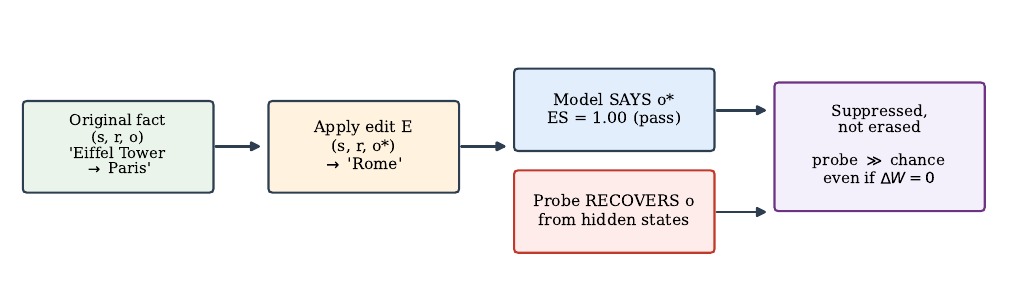}
  \caption{The residual-trace question. A successful edit makes the model \emph{say} the counterfactual object $o^{*}$ (edit success $1.00$), but a linear probe still recovers the \emph{original} object $o$ from the model's hidden states---even for a memory-based editor that changes no base weights ($\Delta W = 0$). We read this as suppression in representational space rather than erasure.}
  \label{fig:pipeline}
\end{figure}

\section{Related Work}
\label{sec:related}

\paragraph{Knowledge editing and its evaluation.}
Locate-then-edit methods modify mid-layer MLP weights identified as carrying factual associations \citep{meng2022locating,meng2023memit}; constrained fine-tuning updates a restricted set of parameters \citep{zhu2020modifying}; memory-based methods such as GRACE \citep{hartvigsen2023aging} route around the base model, storing overrides in an external structure and leaving pretrained weights untouched. Standard evaluation on CounterFact \citep{meng2022locating} measures efficacy, paraphrase, and specificity. A growing audit literature shows these certificates can overstate success: \citet{yang2025mirage} report that generation-based evaluation deflates teacher-forced success rates substantially, and \citet{cohen2024ripple} show single edits fail to propagate to entailed facts. We follow this audit tradition but turn from \emph{whether the new fact is present} to \emph{whether the old fact is still internally decodable}.

\paragraph{Residual knowledge and extraction.}
\citet{patil2024can} recovered ostensibly deleted answers from ROME-edited models using hidden-state probing and rephrasing attacks, establishing that residual information \emph{exists}. \citet{hong2024intrinsic} traced unlearned knowledge in parameter space, and in the unlearning setting relearning attacks restore suppressed capabilities from small corpora \citep{hu2025unlearning,fan2025towards}. Our trace probe is closest to \citet{patil2024can}; we differ by holding the editor's behavioral success fixed at $100\%$, comparing across editors with very different mechanisms, and---most importantly---including a memory-based editor that changes no weights, which lets us ask \emph{where} the residual trace lives rather than only whether it exists.

\paragraph{Mechanisms of editing.}
Mechanistic analyses suggest edits often add suppressive structure rather than removing the original association \citep{xie2025superficial}. Our representational result is consistent with this view and provides a simple, editor-agnostic behavioral-plus-probing measurement of it.

\section{Method}
\label{sec:method}

\paragraph{Setup.}
An edit replaces a true triple $(s, r, o)$---subject, relation, true object, e.g.\ (\textit{Eiffel Tower}, \textit{located-in}, \textit{Paris})---with a counterfactual target $(s, r, o^{*})$, e.g.\ $o^{*}=$\textit{Rome}, using editor $E$ on model $M$, yielding $M_E$. We use GPT-2-XL (1.5B), the canonical editing testbed, loaded in half precision. Edits are drawn from CounterFact \citep{meng2022locating} via the standard tracing split \citep{nanda2022counterfact}; we use the first $N=50$ entries that are well-formed. We study three editors spanning distinct mechanisms:

\begin{itemize}
    \item \textbf{ROME} \citep{meng2022locating} locates a mid-layer MLP as the site of the factual association and applies a closed-form \emph{rank-one} update to its down-projection weight matrix, chosen so a specific key vector (derived from $s$) now maps to a value vector encoding $o^{*}$, while leaving other keys approximately undisturbed. It is a single, non-iterative weight edit.
    \item \textbf{FT-L} \citep{zhu2020modifying} instead treats the same down-projection as a small fine-tuning target: it runs constrained gradient descent (Adam, gradient clipping) on that one weight matrix until the model produces $o^{*}$ for the edit prompt, then stops. Unlike ROME it is iterative and gradient-based, giving a mechanistically distinct route to the same behavioral outcome.
    \item \textbf{GRACE} \citep{hartvigsen2023aging} does not touch base weights at all. It installs a discrete key--value entry in an external adaptor and a forward hook at a chosen layer: at inference, if the incoming activation matches the stored key within a threshold, the hook substitutes a value that steers generation toward $o^{*}$. The base model's parameters are byte-for-byte unchanged; the override is external and reversible by removing the hook.
\end{itemize}

\noindent These three mechanisms let us ask the same residual-trace question (\S\ref{sec:probe}) across a closed-form weight edit, an iterative weight edit, and a no-weight-change external override---so any result that holds across all three is unlikely to be an artifact of one editor's implementation. All edit-success judgments are \emph{generation-based} (greedy decoding, alias matching), never teacher-forced, following \citet{yang2025mirage}.

\paragraph{Weight-change magnitude.}
For each edit we record $\lVert W_{\mathrm{post}} - W_{\mathrm{pre}} \rVert_{1}$ on the targeted down-projection. This is the load-bearing quantity for our mechanism argument: it is large for ROME and FT-L and \emph{exactly zero} for GRACE, which by construction never modifies base weights.

\paragraph{Instrument: representational trace probe.}
\label{sec:probe}
A \emph{linear probe} \citep{alain2017understanding} is a simple classifier---here, logistic regression---trained directly on a model's internal activations to test whether a target quantity is \emph{linearly decodable} from them: if a linear function of the hidden state predicts the target well above chance, that information is represented in a form the rest of the network could in principle read out with a single linear operation. We use it to ask, per edited fact, a strictly binary question: \emph{does the original object $o$ still linearly dominate the counterfactual object $o^{*}$ in the model's representation, or does $o^{*}$ prevail?}

Concretely, after editing we extract the hidden state $h^{(\ell)}(s,r)$ of $M_E$ at the edit prompt, at each candidate layer $\ell$. We compute two scalar projections, $h^{(\ell)}\!\cdot\! u(o)$ and $h^{(\ell)}\!\cdot\! u(o^{*})$, where $u(\cdot)$ is each object's direction in the model's unembedding space, and assign the binary label $y=1$ if $o$'s projection is larger (the original fact still ``wins'' at that layer) and $y=0$ otherwise. The probe is a logistic-regression classifier (standardized features, cross-validated) trained to predict $y$ from $h^{(\ell)}$ across the edited facts in a cell; we report the best-layer cross-validated accuracy \probe{} against an explicit chance baseline of $0.5$. Because $y$ is itself derived from which object's direction the hidden state favors, a probe accuracy above chance means the hidden state carries linearly separable, layer-local signal about which object is still represented as dominant---not merely that \emph{some} residual information about $o$ exists in some nonlinear form. This is a deliberately conservative test: it can only find a trace, never rule one out (\S\ref{sec:limitations}). Hidden states that are numerically non-finite under a given editor are unusable as probe features and are excluded; we report how many are excluded per editor, as this is itself informative (\S\ref{sec:caveats}).

\paragraph{Instrument: resurfacing.}
\label{sec:resurf}
As a behavioral complement we measure the \emph{resurfacing rate} \resurf: the fraction of a small stress set of paraphrase and long-context prompts on which the original object $o$ reappears in generation from $M_E$. \resurf{} is a strict, generation-level signal and we expected it to be sparse; we report it for completeness.

\paragraph{Instrument: relearning savings (reported as a null).}
\label{sec:savings-method}
We also implemented a relearning-savings score \savings{} in the tradition of \citet{ebbinghaus1913memory}: the relative optimization cost to \emph{restore} the original fact versus to teach a matched \emph{novel} fact under equal exposure, with $\savings = 1 - C_{\mathrm{restore}}/C_{\mathrm{novel}}$. The intent was that $\savings>0$ would indicate latent memory. As we report in \S\ref{sec:savings-null}, this instrument did not behave reliably in our setting---especially for the memory-based editor---and we do not draw conclusions from it.

\section{Results}
\label{sec:results}

\paragraph{Edits succeed behaviorally across all three editors.}
All 50 edits reach generation-based success for every editor (ES${}=1.00$; Table~\ref{tab:main}). This is not itself a novel finding---ROME, FT-L, and GRACE are all established methods reporting high success on CounterFact-style edits in prior work, so near-ceiling ES on 50 hand-checked, well-formed edits is expected rather than surprising. We report it because it is load-bearing for everything that follows, not because it is informative on its own: ES$=1.00$ is what licenses the paper's central claim. Without it, an above-chance probe result could trivially reflect an edit that never really took hold behaviorally in the first place---the probe finding the original object simply because the model still says it. By first confirming every edit is behaviorally complete under a strict, non-teacher-forced criterion, we rule out that explanation, so the probe result in Table~\ref{tab:main} can only be read as residual \emph{representational} trace \emph{despite} a genuinely successful behavioral edit, not as a symptom of an incomplete one. The weight-change magnitude cleanly separates the mechanisms: ROME and FT-L make large changes to the targeted weights ($\lVert\Delta W\rVert_1 \approx 1.1\times10^{4}$ and $3.2\times10^{4}$ respectively), while GRACE changes the base weights by \emph{exactly zero}, as designed.

\paragraph{The original object remains decodable after a successful edit.}
The central result is the probe column of Table~\ref{tab:main}. After a successful edit, the original object is linearly recoverable from hidden states well above the $0.50$ chance level for all three editors: $\probe = 0.96$ for ROME, $0.86$ for FT-L, and $0.79$ for GRACE. In other words, a behaviorally successful edit---one that makes the model \emph{say} the new fact---does not remove the original association from the model's internal representation. This is the sense in which the edited fact is suppressed, not erased.

\paragraph{The trace is not an artifact of incomplete weight editing.}
The GRACE row is the key control. GRACE overrides the fact through an external memory and changes no base-model weights ($\lVert\Delta W\rVert_1 = 0$). If the residual probe signal were merely a symptom of a weight edit that did not fully ``reach'' the stored fact, GRACE---which performs no weight edit at all---should expose the original object trivially and uninformatively. Instead GRACE gives an intermediate, clearly above-chance probe ($0.79$), and the dissociation between its zero weight change and the weight editors' large changes shows that the probe is tracking a property of the \emph{underlying representation} that all three editors leave substantially intact, rather than the size of the parameter update. The original association persists in the network whether the editor overwrites weights (ROME, FT-L) or never touches them (GRACE).

\begin{table}[t]
\centering
\small
\begin{tabular}{@{}lccccc@{}}
\toprule
\textbf{Editor} & \textbf{ES} & $\lVert\Delta W\rVert_1$ & \probe & \textbf{Chance} & \resurf \\
\midrule
ROME  & $1.00$ & $1.1\!\times\!10^{4}$ & $\mathbf{0.96}$ & $0.50$ & $0.10$ \\
FT-L  & $1.00$ & $3.2\!\times\!10^{4}$ & $\mathbf{0.86}$ & $0.50$ & $0.01$ \\
GRACE & $1.00$ & $0$ & $\mathbf{0.79}$ & $0.50$ & $0.03$ \\
\bottomrule
\end{tabular}
\caption{Residual-trace audit on GPT-2-XL, $N=50$ CounterFact edits. ES: generation-based edit success (no teacher forcing). $\lVert\Delta W\rVert_1$: $L_1$ change to the targeted down-projection (exactly $0$ for the memory-based editor GRACE). \probe: best-layer cross-validated accuracy of a linear probe recovering the \emph{original} object from hidden states of the edited model, against chance $0.50$. \resurf: generation-level resurfacing rate of the original object under paraphrase/long-context stress. The above-chance probe for all three editors---including weight-free GRACE---is the paper's central result. GRACE's probe is computed on the $14/50$ facts with finite hidden states (\S\ref{sec:caveats}).}
\label{tab:main}
\end{table}

\begin{figure}[t]
  \centering
  \includegraphics[width=0.82\columnwidth]{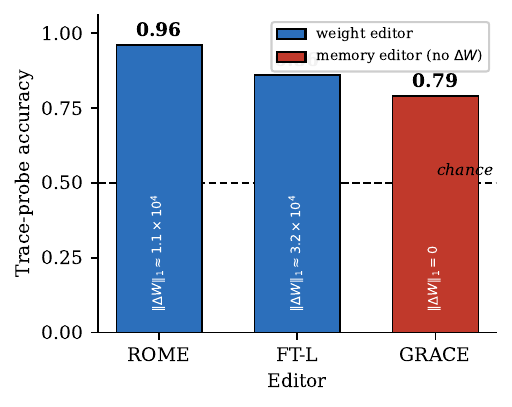}
  \caption{Trace-probe accuracy by editor on GPT-2-XL ($N=50$). All three editors leave the original object decodable well above the chance level (dashed), despite all reaching $100\%$ edit success. The memory-based editor GRACE (red) changes \emph{no} base weights ($\lVert\Delta W\rVert_1=0$) yet still yields an above-chance trace, showing the residual signal is a property of the representation rather than an artifact of an incomplete weight update. GRACE's value is computed on the $14/50$ facts with finite hidden states (\S\ref{sec:caveats}).}
  \label{fig:probe}
\end{figure}

\paragraph{Behavioral resurfacing is sparse.}
The resurfacing rate \resurf{} is low for all editors ($0.01$--$0.10$; Table~\ref{tab:main}): the original object rarely reappears verbatim in greedy generation under simple paraphrase or long-context stress, even though it remains linearly decodable. We read this as a gap between \emph{representational} persistence (which the probe detects) and \emph{behavioral} re-expression, and do not lean on \resurf{} as evidence either way (per-editor caveats in \S\ref{sec:caveats}).

\section{The Savings Instrument: A Negative Result}
\label{sec:savings-null}

We had intended a third instrument, relearning savings (\S\ref{sec:savings-method}), to provide a behavioral-cost measure of latent memory. It did not behave reliably in our setting, and we report this rather than omit it.

Two problems arose. First, the cost-to-criterion is measured at a coarse granularity (the restoration criterion is checked every five optimization steps), so the smallest distinguishable cost is five steps; this is too coarse relative to the effect sizes we sought to detect. Second, and more fundamentally, the instrument is ill-posed for memory-based editing: GRACE's override remains \emph{active} during the restoration attempt, so ``restoring'' the original fact requires optimizing against a live external override that keeps forcing the new answer. The result is that restoration repeatedly hits the step cap, producing large negative savings (mean $\savings \approx -5.2$ for GRACE) that reflect the override fighting relearning, not the absence of a latent trace. For the weight editors the savings values were near zero or mildly negative (ROME $\approx 0.00$, FT-L $\approx -0.30$) and not interpretable at the available granularity.

We therefore do not use savings as evidence. The episode is instructive: a relearning-cost instrument must deactivate any external override before measuring restoration, and must resolve cost more finely than the effect it targets. We leave a corrected savings instrument to future work and rely on the probe, which does not share these failure modes.

\section{Caveats by Editor}
\label{sec:caveats}

Each instrument has editor-specific caveats worth stating explicitly rather than only in aggregate.

\paragraph{GRACE.} The probe value ($0.79$) is computed on a reduced set. GRACE's forward-hook override produces numerically non-finite hidden states on many prompts in our setup; these are unusable as probe features and are excluded. In our run, $36$ of $50$ facts were excluded for GRACE, so its probe rests on $14$ facts, against $50$ for ROME and FT-L. The GRACE probe should therefore be read as suggestive rather than as tightly estimated, and we report the exclusion count explicitly rather than presenting $0.79$ as a clean $N=50$ number. The qualitative conclusion---that an above-chance trace survives even a weight-free edit---does not depend on the precise value, but its precision is limited by this exclusion, and resolving the non-finite-state issue is the first item of future work. GRACE's savings value is also not usable (\S\ref{sec:savings-null}): its override stays active during restoration, which the instrument was not designed to handle.

\paragraph{ROME.} The probe rests on the full $N=50$ with no exclusions, and is the strongest and most cleanly estimated result in the paper ($0.96$). Its main caveat is generality: ROME's rank-one update targets a single layer by construction, so we cannot yet say whether the trace we detect is localized to that layer or distributed, since our probe sweeps candidate layers but the paper does not report a full per-layer profile (left to future work, \S\ref{sec:limitations}).

\paragraph{FT-L.} Also a clean $N=50$ with no exclusions ($0.86$). Its main caveat concerns the savings instrument rather than the probe: because FT-L edits the same weights the savings instrument also fine-tunes during restoration, the two are not fully independent operations on the same parameters, which we believe contributes to its uninterpretable savings value ($\approx-0.30$) alongside the granularity issue described in \S\ref{sec:savings-null}.

\paragraph{Across all three.} Resurfacing (\resurf) is low and noisy for all editors ($0.01$--$0.10$; Table~\ref{tab:main}) and we do not treat it as a reliable signal in either direction; it is reported for completeness only, not as evidence for or against a trace.

\section{Discussion}
\label{sec:discussion}

\paragraph{Editing audits should report what remains.}
If a behaviorally successful edit leaves the original object linearly decodable, then ``the edit succeeded'' is an incomplete certificate. For privacy- and safety-motivated deletion, the quantity that matters is how recoverable the suppressed content is \citep{patil2024can,hu2025unlearning}; a representational trace probe is a cheap, editor-agnostic way to report it alongside efficacy.

\paragraph{Suppression lives in representation, not (only) in weights.}
The GRACE control sharpens the usual reading of residual-knowledge results. Because GRACE changes no weights yet still leaves an above-chance trace, the persistence of the original association is not merely an incompleteness of weight editing; it is a property of the representation that editors operate on top of. This suggests that methods aiming at genuine deletion may need to target the representation directly rather than overwriting or overriding the output mapping.

\section{Conclusion}
\label{sec:conclusion}

On GPT-2-XL, across three mechanistically distinct editors and 50 CounterFact edits, a linear probe recovers the original object from the edited model's hidden states well above chance, even when the edit is fully successful behaviorally and even when the editor---GRACE---changes no base weights at all. Editing, in this setting, suppresses the original fact in representational space rather than erasing it. We release the measurement pipeline, and we report transparently both the strength of the probe result and the failure of a relearning-savings instrument we could not make reliable. Extending the audit to more models and to sequential editing, and repairing the savings and resurfacing instruments, are natural next steps.

\section*{Limitations}
\label{sec:limitations}
This study is deliberately narrow and we state its bounds plainly. (1) \textbf{One model.} All results are on GPT-2-XL; we make no claim about larger or instruction-tuned models, where suppression dynamics may differ. (2) \textbf{One experiment.} We report single-edit residual traces only; we do not study sequential editing, collateral damage, or identity drift. (3) \textbf{Probe lower-bounds information.} A linear probe detects \emph{linearly decodable} signal; above-chance accuracy shows a trace exists, but a lower value would not prove absence, only linear undetectability. (4) \textbf{GRACE probe rests on $14/50$ facts} due to non-finite hidden states under its hook (\S\ref{sec:caveats}); its precise value is uncertain even though the qualitative conclusion is not. (5) \textbf{The savings instrument failed} in our setting and is reported as a null, not a result (\S\ref{sec:savings-null}); the resurfacing signal is sparse and we do not lean on it. (6) \textbf{Scope of editors and data:} three editors, $N=50$ English encyclopedic CounterFact edits; other editors, larger $N$, and other domains are future work. The contribution is the probe-based finding and its weight-free control, not a complete editing audit.

\section*{Ethics Statement}
Our trace probe quantifies how recoverable an edited-away fact remains inside a model. This has dual-use character shared with prior extraction work \citep{patil2024can,hu2025unlearning}: the same measurement that audits a privacy-motivated deletion could in principle guide recovery. We use only public models and public encyclopedic facts; no personal or hazardous information is edited or recovered. We release code oriented toward auditing the completeness of deletion, which we believe is a precondition for honest deletion claims.

\section*{Acknowledgments}
Anonymized for review.

\bibliography{custom.bib}

@inproceedings{meng2022locating,
  title     = {Locating and Editing Factual Associations in {GPT}},
  author    = {Meng, Kevin and Bau, David and Andonian, Alex and Belinkov, Yonatan},
  booktitle = {Advances in Neural Information Processing Systems 35 (NeurIPS)},
  year      = {2022},
  url       = {https://arxiv.org/abs/2202.05262}
}

@inproceedings{meng2023memit,
  title     = {Mass-Editing Memory in a Transformer},
  author    = {Meng, Kevin and Sen Sharma, Arnab and Andonian, Alex and Belinkov, Yonatan and Bau, David},
  booktitle = {The Eleventh International Conference on Learning Representations (ICLR)},
  year      = {2023},
  url       = {https://arxiv.org/abs/2210.07229}
}

@article{zhu2020modifying,
  title   = {Modifying Memories in Transformer Models},
  author  = {Zhu, Chen and Rawat, Ankit Singh and Zaheer, Manzil and Bhojanapalli, Srinadh and Li, Daliang and Yu, Felix and Kumar, Sanjiv},
  journal = {arXiv preprint arXiv:2012.00363},
  year    = {2020},
  url     = {https://arxiv.org/abs/2012.00363}
}

@inproceedings{hartvigsen2023aging,
  title     = {Aging with {GRACE}: Lifelong Model Editing with Discrete Key-Value Adaptors},
  author    = {Hartvigsen, Thomas and Sankaranarayanan, Swami and Palangi, Hamid and Kim, Yoon and Ghassemi, Marzyeh},
  booktitle = {Advances in Neural Information Processing Systems 36 (NeurIPS)},
  year      = {2023},
  url       = {https://arxiv.org/abs/2211.11031}
}

@article{cohen2024ripple,
  title   = {Evaluating the Ripple Effects of Knowledge Editing in Language Models},
  author  = {Cohen, Roi and Biran, Eden and Yoran, Ori and Globerson, Amir and Geva, Mor},
  journal = {Transactions of the Association for Computational Linguistics},
  volume  = {12},
  pages   = {283--298},
  year    = {2024},
  url     = {https://aclanthology.org/2024.tacl-1.16/}
}

@inproceedings{yang2025mirage,
  title     = {The Mirage of Model Editing: Revisiting Evaluation in the Wild},
  author    = {Yang, Wanli and Sun, Fei and Tan, Jiajun and Ma, Xinyu and Cao, Qi and Yin, Dawei and Shen, Huawei and Cheng, Xueqi},
  booktitle = {Proceedings of the 63rd Annual Meeting of the Association for Computational Linguistics (ACL)},
  year      = {2025},
  url       = {https://arxiv.org/abs/2502.11177}
}

@inproceedings{patil2024can,
  title     = {Can Sensitive Information Be Deleted From {LLMs}? Objectives for Defending Against Extraction Attacks},
  author    = {Patil, Vaidehi and Hase, Peter and Bansal, Mohit},
  booktitle = {The Twelfth International Conference on Learning Representations (ICLR)},
  year      = {2024},
  url       = {https://arxiv.org/abs/2309.17410}
}

@inproceedings{hu2025unlearning,
  title     = {Unlearning or Obfuscating? Jogging the Memory of Unlearned {LLMs} via Benign Relearning},
  author    = {Hu, Shengyuan and Fu, Yiwei and Wu, Zhiwei Steven and Smith, Virginia},
  booktitle = {The Thirteenth International Conference on Learning Representations (ICLR)},
  year      = {2025},
  url       = {https://arxiv.org/abs/2406.13356}
}

@article{fan2025towards,
  title   = {Towards {LLM} Unlearning Resilient to Relearning Attacks: A Sharpness-Aware Minimization Perspective and Beyond},
  author  = {Fan, Chongyu and Jia, Jinghan and Zhang, Yihua and Ramakrishna, Anil and Hong, Mingyi and Liu, Sijia},
  journal = {arXiv preprint arXiv:2502.05374},
  year    = {2025},
  url     = {https://arxiv.org/abs/2502.05374}
}

@article{hong2024intrinsic,
  title   = {Intrinsic Evaluation of Unlearning Using Parametric Knowledge Traces},
  author  = {Hong, Yihuai and Yu, Lei and Ravfogel, Shauli and Yang, Haiqin and Geva, Mor},
  journal = {arXiv preprint arXiv:2406.11614},
  year    = {2024},
  url     = {https://arxiv.org/abs/2406.11614}
}

@inproceedings{xie2025superficial,
  title     = {Revealing the Deceptiveness of Knowledge Editing: A Mechanistic Analysis of Superficial Editing},
  author    = {Xie, Jiakuan and Cao, Pengfei and Chen, Yuheng and Chen, Yubo and Liu, Kang and Zhao, Jun},
  booktitle = {Proceedings of the 63rd Annual Meeting of the Association for Computational Linguistics (ACL)},
  year      = {2025}
}

@inproceedings{geva2021transformer,
  title     = {Transformer Feed-Forward Layers Are Key-Value Memories},
  author    = {Geva, Mor and Schuster, Roei and Berant, Jonathan and Levy, Omer},
  booktitle = {Proceedings of the 2021 Conference on Empirical Methods in Natural Language Processing (EMNLP)},
  year      = {2021},
  url       = {https://aclanthology.org/2021.emnlp-main.446/}
}

@inproceedings{hernandez2024linearity,
  title     = {Linearity of Relation Decoding in Transformer Language Models},
  author    = {Hernandez, Evan and Sen Sharma, Arnab and Haklay, Tal and Meng, Kevin and Wattenberg, Martin and Andreas, Jacob and Belinkov, Yonatan and Bau, David},
  booktitle = {The Twelfth International Conference on Learning Representations (ICLR)},
  year      = {2024},
  url       = {https://arxiv.org/abs/2308.09124}
}

@book{ebbinghaus1913memory,
  title     = {Memory: A Contribution to Experimental Psychology},
  author    = {Ebbinghaus, Hermann},
  publisher = {Teachers College, Columbia University},
  address   = {New York},
  year      = {1913},
  note      = {Translated by H.~A.~Ruger and C.~E.~Bussenius; original work published 1885}
}

@misc{nanda2022counterfact,
  title={{CounterFact}-Tracing Dataset},
  author={Nanda, Neel},
  year={2022},
  note={HuggingFace dataset: NeelNanda/counterfact-tracing, adapted from \citet{meng2022locating}}
}

@inproceedings{alain2017understanding,
  title={Understanding Intermediate Layers Using Linear Classifier Probes},
  author={Alain, Guillaume and Bengio, Yoshua},
  booktitle={International Conference on Learning Representations (ICLR) Workshop},
  year={2017}
}

\appendix

\section{Implementation Details}
\label{sec:impl}
Edits use a fixed mid-layer down-projection as the edit site. ROME applies a rank-one update derived from a real intermediate activation; FT-L fine-tunes the down-projection with constrained gradient steps (gradient clipping; the trained weights are upcast to full precision during the update for numerical stability under half-precision inference); GRACE installs an override through a forward hook and modifies no base weights. The probe is logistic regression with standardized features and cross-validation, trained per candidate layer on hidden states extracted at the edit prompt; we report the best-layer cross-validated accuracy. Edit success is greedy-decoding generation with alias matching. The pipeline is fully resumable via an append-only ledger so that runs can be checkpointed and reproduced. All code and the result ledger are released.

\end{document}